\documentclass[runningheads,hidelinks]{llncs}
\usepackage[T1]{fontenc}

\usepackage{lmodern}
\usepackage{amsmath}
\usepackage{amsfonts}

\usepackage[english]{babel} %
\usepackage[babel]{csquotes} %
\usepackage[breaklinks,pdftitle=Persistence\ Initialization,pdfdisplaydoctitle]{hyperref}
\usepackage{booktabs}
\usepackage{pgf}
\usepackage{bm}
\usepackage{color}
\usepackage{wrapfig}
\usepackage{graphicx}
\usepackage{etoolbox}
\apptocmd{\sloppy}{\hbadness 10000\relax}{}{}

\usepackage[flushleft]{threeparttable}

\DeclareMathOperator{\smape}{s\textsc{mape}}
\DeclareMathOperator{\mase}{\textsc{mase}}
\DeclareMathOperator{\owa}{\textsc{owa}}

\begin{document}

\title{Persistence Initialization:\texorpdfstring{\\}{ }A novel adaptation of the Transformer architecture for\texorpdfstring{\\}{ }Time Series Forecasting}
\titlerunning{Persistence Initialization}
\author{Espen Haugsdal, Erlend Aune, Massimiliano Ruocco}
\institute{Norwegian University of Science and Technology}

\maketitle

\begin{abstract}
  Time series forecasting is an important problem, with many real world applications.
  Ensembles of deep neural networks have recently achieved impressive forecasting accuracy, but such large ensembles are impractical in many real world settings.
  Transformer models been successfully applied to a diverse set of challenging problems.
  We propose a novel adaptation of the original Transformer architecture focusing on the task of time series forecasting, called Persistence Initialization.
  The model is initialized as a naive persistence model by using a multiplicative gating mechanism combined with a residual skip connection.
  We use a decoder Transformer with ReZero normalization and Rotary positional encodings, but the adaptation is applicable to any auto-regressive neural network model.
  We evaluate our proposed architecture on the challenging M4 dataset, achieving competitive performance compared to ensemble based methods.
  We also compare against existing recently proposed Transformer models for time series forecasting, showing superior performance on the M4 dataset.
  Extensive ablation studies show that Persistence Initialization leads to better performance and faster convergence.
  As the size of the model increases, only the models with our proposed adaptation gain in performance.
  We also perform an additional ablation study to determine the importance of the choice of normalization and positional encoding, and find both the use of Rotary encodings and ReZero normalization to be essential for good forecasting performance.
  \keywords{Transformer \and Time Series Forecasting \and M4 Competition}
\end{abstract}

\section{Introduction}
Being able to accurately forecast the future can allow for better decision-making in the present, making forecasting a very valuable tool across a wide range of fields. Examples include finance, econometrics, internet of things and preventive maintenance.

Much of the success of deep neural networks has been attributed to their ability to learn general representations, so-called representation learning.
In the context of time series forecasting, the importance of learning such general representations has recently been validated by the high forecasting accuracy of deep neural networks~\cite{oreshkin2019n,smyl2020hybrid} on the challenging M4 forecasting dataset \cite{makridakis2020m4}.
The M4 dataset is a large collection of 100,000 time series, and currently the most accurate methods are ensembles of deep neural networks, which learn shared representations from multiple time series.
This is in contrast to the approach of most ``classical'' statistical learning methods for time series, such as ARIMA, which consider each time series to be a single independent learning problem.
However, the reliance on ensembles indicates that the deep neural networks were not able to represent many of the features of the data in a single model.

Ensembles are often used in competitions, however in many real world applications they are impractical due to deployment complexity or model size.
This was famously the case in the Netflix Competition, where the second place entry was preferred due to the complexity of the first place entry, which was a large ensemble.

The Transformer~\cite{vaswani2017attention} has been successfully applied to a diverse set of challenging problems.
The ability of the Transformer to learn complex relationships makes it a promising alternative to ensemble based methods.
However, the use of Transformers for time series forecasting has received relatively little attention compared to other domains.

In this work, we present an adaptation which significantly improves the forecasting accuracy of the Transformer architecture.
In summary, our contributions are as follows:
\begin{enumerate}
  \item We present an adaptation of the Transformer architecture to make it more suitable for time series forecasting.
        This adaptation initializes the model such that it starts off as a naive random walk model, which provides a good starting point for further learning. %
        The random walk initialization can be achieved by combining a zero-initialized multiplicative gating mechanism with a residual skip connection.
        We perform an ablation study to verify the importance of both the skip connection and the gating mechanism.
  \item We show that a single Transformer model, with the proposed adaptation, can achieve competitive forecasting performance on the challenging M4 dataset when compared to large ensembles of deep networks.
  \item We perform an ablation study on the choice of positional encoding and normalization layers.
        We show that both the positional encoding and the normalization layer has a large effect on the performance of the model.
        Using a relative positional encoding (the Rotary encoding~\cite{su2021roformer}), leads to better forecasting performance compared to the standard sinusoidal encoding.
        Using a ReZero Transformer leads to performance improving with increasing model size, while using normalization layers leads to worse performance the model increases.
  \item We perform a comparison with recent existing Transformer architectures for the task of time series forecasting. Results show superior performance of our proposed architecture on the well established M4 benchmark dataset for time series forecasting.
\end{enumerate}
The rest of the paper is organized as follows:
Section \ref{sec:background} reviews existing related work, Section \ref{sec:methods} describes our proposed architecture, Section \ref{sec:experimets} describes the experiments, and Section \ref{sec:results} provides analysis and discussion of the results.
Finally, Section \ref{sec:conclusion} concludes with a summary.

\section{Background and Related Work}
\label{sec:background}
\subsubsection{Transformers for time series forecasting.}
Li et al. \cite{li2019enhancing} were among the first to focus on using a Transformer for time series forecasting.
The authors introduced the LogSparse Transformer, which uses a sparse attention operation by exponentially increasing the space between past consecutive queries, achieving $N(\log N)^2$ complexity.
Furthermore, the authors used a causal convolution operation to make the attention operation aware of a range of points.
The model outperformed previous RNN based methods in forecasting electricity and traffic data.

Informer \cite{informer} and Autoformer \cite{wu2021autoformer} both focus on the task of Long Sequence Time-Series Forecasting, which they define as forecasting horizons of size 48 or longer.
Both methods produce forecasts for the entire horizon in a single evaluation, improving inference speed compared to an auto-regressive Transformer.
Moreover, both works propose efficient attention mechanisms in order to avoid the issue of quadratic complexity scaling.
The Informer only uses the dominant queries in the attention computation, based on an approximation of the query-key similarity.
The Autoformer instead replaces the entire dot-product attention mechanism with an auto-correlation based mechanism.
The Informer was evaluated on long range electricity and weather data. %
The Autoformer was also evaluated on long range data, with extensive testing in multiple domains: electricity, currency exchange, traffic, weather, disease.
Both methods outperformed prior work, improving the state of the art in long range forecasting.

Zerveas et al. \cite{zerveas2021transformer} focus on time series classification and regression, but do unsupervised pre-training by using a masked autoencoder objective.
The authors evaluated their approach using datasets from the UEA and UCR time series repositories.

In contrast to these works, we evaluate our approach on the M4 dataset\cite{makridakis2020m4}, a challenging and well established forecasting dataset in the time series forecasting community.
To the best of our knowledge, our work is the first Transformer to achieve competitive results on the complete M4 dataset.

\subsubsection{Skip connections and gating mechanisms.}
Residual skip connections have become ubiquitous in neural network architecture design, due to their stabilizing effect when training deep networks.
ResNet \cite{he2016deep} is the most famous example of such skip connections.
However, the simple skip connection in ResNet can be seen as a special case of Highway Networks~\cite{srivastava2015highway}, which also includes multiplicative gating.

Normalization layers such as Batch Normalization~\cite{ioffe2015batch} and Layer Normalization~\cite{ba2016layer} has also played an important role in training deeper networks.
However, even with normalization layers, deep Transformer models can be difficult to train. %
ReZero \cite{bachlechner2021rezero} proposed a simple alternative to Layer Normalization, by instead using a residual connection and a scalar gating parameter initialized to 0 after each layer, which effectively initializes every layer to perform the identity operation.
This ReZero Transformer achieved 56\% faster convergence than a regular Transformer on a benchmark NLP task.
Moreover, the authors are able to train Transformers with more than 100 layers, without the need for normalization, auxiliary losses, or learning rate warm up.

Our proposed Persistence Initialization adaptation also consists of a residual skip connection and a learnable scalar gating mechanism, similarly to ReZero.
However, Persistence Initialization is not a technique to increase network depth, but rather a reparametrization of the learning problem.

\subsubsection{The M4 competition.}
The Makridakis competitions have been important for the development of forecasting methods.
The fourth Makridakis competition \cite{makridakis2020m4}, known has the M4 competition, was held in 2018 and featured a dataset containing 100000 time series of various lengths.
The time series were divided frequencies, each with its pre-defined forecasting horizon $H$: Yearly~($H=6$), Quarterly~($H=8$), Monthly~($H=18$), Weekly~($H=13$), Daily~($H=14$), and Hourly~($H=48$).

The winner of the competition, Smyl \cite{smyl2020hybrid}, used a hybrid model which combined a dilated-attention LSTM with several exponential smoothing models learned independently for each time series.
Multiple such models were used in an complex hierarchical ensemble to produce the final forecast:
First, 6-9 independent models were used.
Second, each data frequency was split into subsets, and multiple models were trained on each subset, if computationally feasible.
Third, the weights of the last few epochs of training were used to produce multiple forecasts.

The second place entry, by Montero-Manso et al. \cite{montero2020fforma}, was a weighted average of nine different statistical methods.
The weights were computed dynamically by using the gradient tree boosting method XGBoost \cite{chen2016xgboost}, and the model was trained on 42 different statistical time series features.

After the conclusion of the M4 competition, Oreshkin et al. \cite{oreshkin2019n} introduced a purely neural network based model, called N\nobreakdash-BEATS, which was able to produce even better forecasts, outperforming the winner of the competition by a substantial margin.
The model was an ensemble of 180 deep MLPs, with each MLP consisting of multiple blocks of feed-forward networks.
Each block produces both a forecast and a backcast, where the forecast of each block is added to the total forecast, and the backcast is subtracted from the input of the following block.
The backcast is used to remove the parts of the input that it can approximate well, while the forecast can be seen as forming a sequence of consecutive partial estimates.
To ensure diversity in the ensemble, the authors used three different loss functions and six different window sizes, resulting in a total of 18 different model configurations.

These models were able to achieve high accuracy, but they were also highly complex, making them difficult to adapt to real world settings.
We propose instead to use a single Transformer model, with a simple training setup.

\section{Methods}
\label{sec:methods}
We define a time series to be a sequence of fully observable measurements \linebreak $\bm{x}=[x_1, \ldots, x_T] \in \mathbb{R}^T $, where $x_t$ is the observation at time $t$ and $T$ is the length of the series.
For a given forecasting horizon $H$, the goal of forecasting is to predict $\bm{y}=[x_{T+1}, \ldots, x_{T+H}] \in \mathbb{R}^H $ from $\bm{x}$.

A sliding window approach is used in order to simplify the learning problem and improve computational efficiency.
In other words, only the $nH$ most recent values of $\bm{x}$ are used to produce the forecast $\bm{\hat{y}}$, where $n$ is a hyper-parameter determining the size of the input window relative to the forecasting horizon.

We normalize the inputs of the model by dividing by the mean of the $H$ most recent values, followed by a log transform:
\begin{equation}
  \bm{z} = f(\bm{x}) = \ln \frac{\bm{x}}{\mu_H(\bm{x})} ,
\end{equation}
where $\mu_H$ computes the mean of the $H$ most recent values within the input portion of the series.
The $H$ most recent values are used in order to better capture the trend of the series close to the forecasting window.
The log transform is used to enforce positive outputs, as all of the values in the M4 dataset are positive.
To produce an output in the data space, we perform the inverse transformation.
During training gradients are propagated through this inverse transformation.

The values of the normalized univariate time series are converted to a feature vector by projecting it to a space of size $d_\text{model}$.
A Transformer is then applied to this sequence of feature vectors.
Similarly, we use another projection to convert the final output of the Transformer back to a univariate series:
\begin{equation}
  g(\bm{z}) = \text{Transformer}(\bm{z} W_\text{in}) W_\text{out}, \label{eq:proj}
\end{equation}
where $W_\text{in} \in \mathbb{R}^{1 \times d_\text{model}}$, and $W_\text{out} \in \mathbb{R}^{d_\text{model} \times 1} $.

\subsection{Persistence Initialization}
\begin{figure}[!t]
  \includegraphics[width=\textwidth]{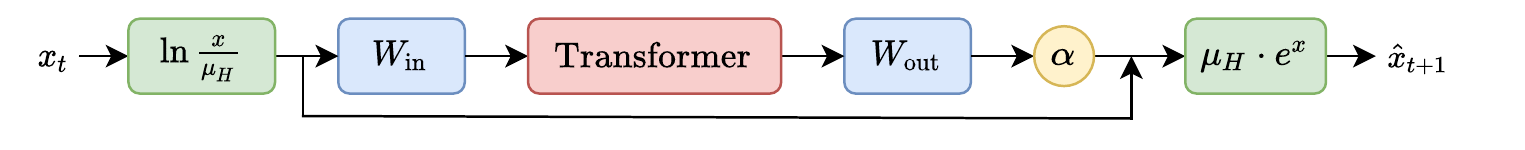}
  \vspace{-20pt}
  \caption{
    The proposed adaptation consists of a skip connection and a scalar multiplicative gating mechanism initialized to 0.
    The initial model becomes the naive persistence model, i.e. the model that predicts $\hat{x}_{t+1} = x_t$.
  }
  \label{fig:arch}
\end{figure}

We propose an adaptation of the original Transformer, in order to make it more suitable for time series forecasting.
A decoder-only architecture is used in order to generate forecasts autoregressively, similarly to generative language models (e.g. GPT-2 \cite{radford2019language}).

The proposed Persistence Initialization is implemented by adding a skip connection and a learnable scalar gating parameter $\alpha$ to the model: %
\begin{equation}
  h(\bm{z})  = \bm{z} + \alpha \cdot g(\bm{z}) \label{eq:skip_alpha}
\end{equation}
The skip connection makes the model to learn the residual function, which is equivalent to learning to predict the differences between consecutive time steps.
Setting the initial $\alpha$ to zero makes the initial model equivalent to a naive persistence model, which simply uses the value at time $t$  as the forecast for $t+1$.
A graphical representation of the adaptation is given in Figure~\ref{fig:arch}.

\subsubsection{Model architecture.}
The decoder-only Transformer architecture consists of causal self-attention layers followed by feedforward layers.
We connect these layers with skip connections using ReZero gating.
The model can be defined as follows:
\begin{align*}
  \text{Transformer}(X) & = X_{N}                                      \\ %
  X_{l}                 & = \text{FF} ( \text{SA} (X_{l-1}))           \\ %
  \text{SA}(X)          & = X + \alpha_l \cdot \text{SelfAttention}(X) \\
  \text{FF}(X)          & = X + \alpha_l \cdot \text{FeedForward}(X),  %
\end{align*}
where $N$ is the number of layers, and $\alpha_l$ is a learnable scalar parameter shared between the self-attention and feedforward layer within each layer.
The feedforward layer is defined as in the original Transformer, consisting of two affine layers with a ReLU activation function in between:
\begin{align*}
  \text{FeedForward}(X) & = \text{ReLU}(XW_1 + b_1) W_2 + b_2,
\end{align*}
where $W_1 \in \mathbb{R}^{d_\text{model} \times d_\text{ff}}$, and $W_2 \in \mathbb{R}^{d_\text{ff} \times d_\text{model}}$.

We use relative positional encodings, or more specifically Rotary encodings~\cite{su2021roformer}, instead of the standard sinusoidal positional encoding.
In the context of windowed time series, encoding the absolute position (i.e. the distance to start of the input window) does not have any particular semantic significance.
In contrast, for text data, the start of the sequence often has a semantic meaning, for instance as the start of a document or a sentence.
Instead of adding a positional encoding to the input features of the model, the Rotary encoding is implemented by modifying the definition of self-attention:
\begin{align*}
  \text{SelfAttention}(X)            & = \text{MultiHeadAttention}(X, X, X)                                                              \\
  \text{MultiHeadAttention}(Q, K, V) & = \text{Concat}(\text{head}_1, ..., \text{head}_h)W^O                                             \\
  \text{head}_i                      & = \text{Attention}(QW_i^Q, KW_i^K, VW_i^V)                                              \nonumber \\
  \text{Attention}(Q,\ K,\ V)        & = \text{softmax}(\frac{\widetilde{Q}\widetilde{K}^T}{\sqrt{d_{qk}}})V,                            %
\end{align*}
where $\widetilde{Q}$ and $\widetilde{K}$ represents the $Q$ and $K$ matrices with Rotary positional encoding applied.
The effect of the encoding is to multiply each key and query with a rotation matrix, which causes positional information to be encoded in the angle between the vectors.

\section{Experiments}
\label{sec:experimets}
\subsection{Experimental setting}
\subsubsection{Dataset.}
We evaluate the performance of our adaptation on the M4 dataset.
The M4 dataset contains 100000 time series of various lengths, divided into six frequencies: yearly, quarterly, monthly, weekly, daily, and hourly.
Table \ref{tab:M4_stats} contains some descriptive statistics for each frequency.
\begin{table}[t!]
  \centering
  \begin{threeparttable}
    \caption{Descriptive statistics for each frequency in the M4 dataset.}
    \label{tab:M4_stats}
    \begin{tabular}{lcccccc}
      \toprule
      {}                       & Yearly & Quarterly & Monthly & Weekly & Daily & Hourly \\
      \midrule
      Number of time series    & 23,000 & 24,000    & 48,000  & 359    & 4227  & 414    \\
      Forecasting horizon, $H$ & 6      & 8         & 18      & 13     & 14    & 48     \\
      Seasonality, $S$         & 1      & 4         & 12      & 1      & 1     & 24     \\
      \midrule
      Minimum length           & 19     & 24        & 60      & 93     & 107   & 748    \\
      25\% percentile length   & 26     & 70        & 100     & 392    & 337   & 748    \\
      50\% percentile length   & 35     & 96        & 220     & 947    & 2954  & 1008   \\
      75\% percentile length   & 46     & 123       & 324     & 1616   & 4211  & 1008   \\
      Maximum length           & 841    & 874       & 2812    & 2610   & 9933  & 1008   \\
      \bottomrule
    \end{tabular}
  \end{threeparttable}
\end{table}

\subsubsection{Metrics.}
The M4 competition introduced a metric called Overall Weighted Average (OWA)~\cite{makridakis2020m4}.
It is defined as a combination of the Mean Absolute Scaled Error (MASE) and symmetric Mean Absolute Percentage Error (sMAPE):
\begin{align}
  \smape & = \frac{1}{N} \sum_{i=1}^N \frac{200}{H} \sum_{j=1}^H \frac{|y_{j}^{(i)} - \hat{y}_{j}^{(i)} |}{|y_{j}^{(i)} | + |\hat{y}_{j}^{(i)} |}                                                                  \\
  \mase  & = \frac{1}{N} \sum_{i=1}^N \frac{ \frac{1}{H} \sum_{j=1}^H  |y_{j}^{(i)} - \hat{y}_{j}^{(i)} |}{\frac{1}{T^{(i)}-S^{(i)}}\sum_{j=S^{(i)}+1}^{T^{(i)}}|x_j^{(i)} - x_{j-S^{(i)}}^{(i)}|} \label{eq:mase} \\
  \owa   & = \frac{1}{2} \left[ \frac{\smape}{\smape_{\textrm{Naïve2}}}  + \frac{\mase}{\mase_{\textrm{Naïve2}}}  \right],
\end{align}
$S^{(i)}$ is defined to be the seasonality for time series $i$, which the M4 competition defined to be a constant for all time series within a data frequency (e.g. 24 for all hourly time series, 12 for all monthly time series).
sMAPE and MASE are both scale independent metrics commonly used in the forecasting literature \cite{hyndman2006another}.
MASE is a scaled version of Mean Absolute Error (MAE), with the scaling factor for a given time series being the MAE of a baseline model.
For the M4 competition, the chosen baseline model was the seasonal naïve model, which always predicts the value $S^{(i)}$ steps in the past.
Finally, in order to combine the sMAPE and MASE scores into a single metric, the scores are normalized by dividing by the corresponding scores from another model, called Naïve2.
The Naïve2 baseline is a persistence model which is seasonally adjusted by multiplicative decomposition if auto-correlation is more than 90\%.

\subsubsection{Hyper-parameters.}
The window size is defined to be $nH$, where $H$ is the forecasting horizon.
For the Yearly, Quarterly, Monthly, and Daily frequencies we set $n=3$, and for the Weekly and Hourly we set $n=4$.
We use a model with 4 layers, and the number of attention heads is set to 4.
$d_\text{model}$ is set to 512, and $d_\text{ff}$ is set to 2048.
We use the Lamb~\cite{you2019large} optimizer with default hyper-parameters.
The bias corrected version of Lamb was used, with gradient clipping for gradient norm greater than 10.
We use the MASE metric (as defined in Equation \ref{eq:mase}) as the loss function.
Early stopping with a patience of 8 epochs was used as as the stopping criterion.
We define a training epoch to consist of 128 mini-batches of size 1024.

The hyper-parameter settings were found using manual search, focusing almost exclusively on the Monthly frequency.
The goal of the search was to find a general setting which could work well across all frequencies.
We were largely successful in this goal, except for the value of $n$, as we found that a value of 3 lead to poor performance on the Weekly and  Hourly frequencies, compared to a value of 4.
We believe this is due to seasonal patterns that are only included with a window size corresponding to $n=4$.
In the case of Weekly, $n=4$ corresponds to 52 weeks, which indicates the presence of yearly seasonality.
For the Hourly frequency, n=4 corresponds to 192 hours, which is a approximately 8 days, indicating a weekly seasonality.

\subsubsection{Training.}
We sample sub-sequences of length $nH + H$, where the first $nH$ elements are treated as inputs, and the final $H$ elements are treated as targets to be learned.
The validation set is constructed by taking the rightmost such sub-sequence.
The training set is created by enumerating all possible sub-sequences without overlapping targets in the validation set.
In order to have a greater number of sub-sequences available for the training set, the shortest sequences (defined as those with length below the 25th percentile) are excluded from the validation set.
To construct a training mini-batch, we first a sample time series with uniform probability, and then sample from the sub-sequences within that time series with (conditional) uniform probability.
Teacher forcing is used to produce $H$ predictions in parallel during training.
During testing the output of the model is produced autoregressively.

\subsection{Ablation studies}
In order to better understand the effects of the various components of our models, we perform two ablation studies.
The first ablation study focuses on the effects of the proposed Persistence Initialization adaptation, and the second study focuses on the effects of the positional encoding and the normalization layers.
We focus only on the Monthly portion of M4, as it is the frequency with the most time series, contributing to 48\% of the overall dataset and total OWA score.
We vary the model size by setting the hyper-parameter $d_\text{model}$ to values in the set $\{ 32, 64, 128, 256, 512 \}$, with the feedforward size $d_\text{ff}$ set to~$4\cdot d_\text{model}$.
For each combination of size setting and ablation variable setting, we train 9 models with different random weight initializations.

\subsubsection{Skip connection and multiplicative gating.}
In this ablation study, we investigate the effects of the skip connection and the multiplicative gating, by modifying Equation~\ref{eq:skip_alpha}.
We compare an architecture with no skip connection and no multiplicative gating (Equation~\ref{eq:skip_alpha_c}), an architecture with a skip connection and no multiplicative gating (Equation~\ref{eq:skip_alpha_b}), and the full architecture with both a skip connection and multiplicative gating (Equation~\ref{eq:skip_alpha_a}):
\begin{align}
  h(\bm{z}) & = g(\bm{z}) \label{eq:skip_alpha_c}                       \\
  h(\bm{z}) & = \bm{z} + g(\bm{z}) \label{eq:skip_alpha_b}              \\
  h(\bm{z}) & = \bm{z} + \alpha \cdot g(\bm{z}) \label{eq:skip_alpha_a}
\end{align}

\subsubsection{Positional encoding and normalization.}
We compare the effect of the positional encoding and normalization layers.
Two positional encodings are compared: the rotary encoding, and the original sinusoidal encoding.
The sinusoidal positional encoding is implemented as a matrix, as follows:
\begin{equation*}
  E_{i,j} =
  \begin{cases}
    \sin{i / 10000^{j/d_\text{model}}}       & j \text{ even}   \\
    \cos{i / 10000^{(j - 1)/d_\text{model}}} & j \text{ odd}\ ,
  \end{cases}
\end{equation*}
\begin{equation*}
  g(\bm{z}) = \text{Transformer}(\bm{z} W_\text{in} + E) W_\text{out}
\end{equation*}

We compare three kinds of normalization: ReZero, post-activation Layer Norm, and pre-activation Layer Norm:
\begin{align}
  \text{ReZero}(x) & = x + \alpha_l \cdot \text{Sublayer}(x)    \\
  \text{PostLN}(x) & = \text{LayerNorm}(x + \text{Sublayer}(x)) \\
  \text{PreLN}(x)  & = x + \text{Sublayer}(\text{LayerNorm}(x))
\end{align}

\subsection{Comparison with other Transformer models for Time Series.}
We perform an experiment to compare the performance of our architecture with other Transformer models which have been applied to time series forecasting: LogSparse Transformer~\cite{li2019enhancing}, Informer~\cite{informer}, and Autoformer~\cite{wu2021autoformer}.
In this experiment we focus on the Hourly frequency to evaluate the models, because it's relatively long forecasting horizon of 48 hours, making it more similar to the long forecasting horizons the Informer and Autoformer were designed for.
For the LogSparse Transformer, we use the authors' own reported performance on the hourly portion data of M4.
For the two other models, we use a similar approach as in previous experiments, and report the median performance of 5 repeated runs.

We used publicly available code\footnote{\url{https://github.com/zhouhaoyi/Informer2020}}\footnote{\url{https://github.com/thuml/Autoformer}} to implement the Informer and Autoformer.
However, we found training the Autoformer to be challenging, as the loss values were generally high throughout training.
This was especially the case as the number of parameters increased, and for this reason we decided to only consider $d_\text{model}= 32$.
Similarly to the previous experiments, we set $d_\text{ff}= 4 \cdot d_\text{model}$.
For the Autoformer and the Informer we use 2 encoder layers and 2 decoder layers, and for the Transformer we use 4 layers, as before.
This setting results in a similar number of parameters for the three models.
We use a context window of length $H$ for the Informer and Autoformer decoder.

We also found the previously used strategy of early stopping on the validation loss to be unreliable when training the Informer and Autoformer, as the validation loss would often have much larger variance than in the previous experiments.
We believe this difference comes from the one-shot forecasting approach taken by both methods.
In contrast, our auto-regressive model uses teacher forcing during training, leading to more stable loss values during training, as the model only has to perform 1-step predictions.
To provide a more fair comparison, we instead allocate a fixed amount of computation to each method by setting a limit of 100 epochs, and then selecting the weights with lowest validation loss to compute the final test score.

The authors of LogSparse Transformer measure performance using a 0.5-quantile loss, so to be comparable with we also report the $R_{0.5}$ quantile loss, which can be defined as follows:
\begin{equation}
  R_{0.5} = \frac{ \sum_{i=1}^N  \sum_{j=1}^H |y_{j}^{(i)} - \hat{y}_{j}^{(i)}|}{\sum_{i=1}^N  \sum_{j=1}^H |y_{j}^{(i)}|} \label{eq:normalized_deviation}
\end{equation}

\section{Results and Discussion}
\label{sec:results}
\begin{table}[t!]
  \centering

  \begin{threeparttable}
    \caption{
      OWA test scores for each frequency of the M4 dataset.
      Bold font is used to indicate the best model, an underline indicates second best.
    }
    \label{tab:M4_OWA}
    \begin{tabular}{lccccccc|c}
      \toprule
      {}                                   & Yearly                  & Quarterly               & Monthly                    & Weekly            & Daily             & Hourly            & Total                   & Ensemble     \\
                                           & (23k)                   & (24k)                   & (48k)                      & (359)             & (4227)            & (414)             & (100k)                  & size         \\
      \midrule
      M4 Rank 1                            & 0.778                   & 0.847                   & 0.836                      & 0.851             & 1.046             & 0.444             & 0.821                   & 6 to 9       \\
      M4 Rank 2                            & 0.799                   & 0.847                   & 0.858                      & 0.796             & 1.019             & 0.484             & 0.838                   & 9            \\
      \midrule
      \textit{N-BEATS-18}\hphantom{0}      & -                       & -                       & -                          & -                 & -                 & -                 & \textit{0.802}          & \textit{18}  \\
      \textit{N-BEATS-180}                 & \textit{\textbf{0.758}} & \textit{\textbf{0.800}} & \textit{\underline{0.819}} & -                 & -                 & -                 & \textit{\textbf{0.795}} & \textit{180} \\
      \midrule
      \textbf{PI-Transformer}\tnote{\dag}  & 0.777                   & 0.852                   & 0.833                      & \underline{0.733} & \textbf{0.987}    & \underline{0.431} & 0.815                   & 1            \\
      \textbf{PI-Transformer}\tnote{\ddag} & \underline{0.769}       & \underline{0.836}       & \textbf{0.813}             & \textbf{0.697}    & \underline{0.987} & \textbf{0.397}    & \underline{0.800}       & 9            \\
      \bottomrule
    \end{tabular}
    \begin{tablenotes}
      \item[\dag] Median OWA scores from 9 repeated experiments. %
      \item[\ddag] Mean ensemble of the predictions of the 9 repeated experiments.
    \end{tablenotes}
  \end{threeparttable}
\end{table}

\subsection{M4 comparison}

In Table \ref{tab:M4_OWA} we compare the forecasting performance of Transformer models trained with our proposed adaptation (\textbf{PI-Transformer} for short) against other state of the art methods.
M4 Rank 1 is the hybrid Exponential Smoothing~/~LSTM model by Smyl~\cite{smyl2020hybrid}, M4 Rank 2 refers to FFORMA model by Montero-Manso et al. \cite{montero2020fforma}, and N-BEATS is the stacked MLP model proposed by Oreshkin et al. \cite{oreshkin2019n}.
All three of these models use ensembles to improve their forecasts.
FFORMA  uses XGBoost to combine the forecasts of nine different statistical methods.
Smyl's hybrid model uses a complex multi-level ensembling approach.
N-BEATS  takes the median forecast of a diverse set of models which are trained using three different loss functions and six different window lengths.

In order to provide robust measurements of the performance of our model, we train 9 models for each frequency and report the median OWA within each frequency.
As can be seen from the table, the median model has a lower total OWA than the winner of the M4 competition, and outperforms the M4 winner in every frequency except for Quarterly.

We are mainly interested in achieving competitive forecasting performance without the need for ensembles.
However, comparing the performance of a single model against a ensemble is not a fair comparison.
For this reason, we provide an estimate of the effect of using our Transformer in an ensemble, by we computing the mean prediction of our 9 models.
This improves performance in every frequency except Daily, which demonstrates that even simple ensembling techniques can have a powerful effect on forecasting performance on the M4 dataset.
Remarkably, the total OWA for the mean prediction is comparable to the version of N-BEATS with 180 models, and the mean prediction outperforms the version of N-BEATS with 18 models.

\begin{figure} %
  \centering
  \input{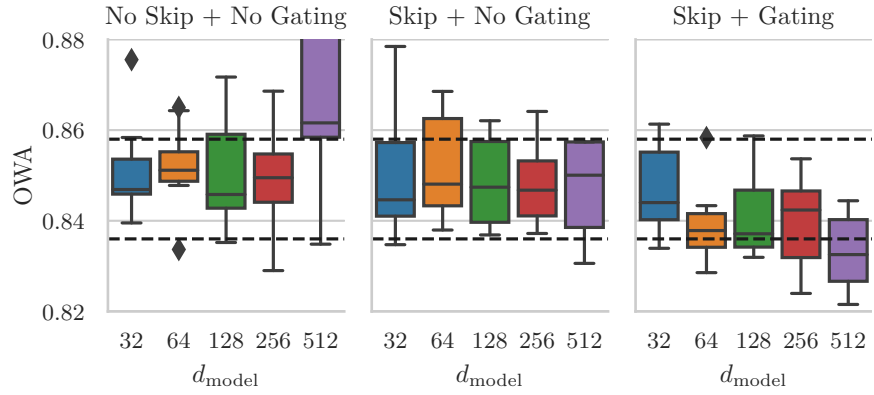}
  \caption{
    Boxplots of OWA scores on the M4-Monthly dataset.
    The dashed lines correspond to the first and second place entry in the M4 competition.
    Each box represents 9 repeated runs.
  }
  \label{fig:skip_owa_boxplot}
\end{figure}
\begin{figure}%
  \centering
  \input{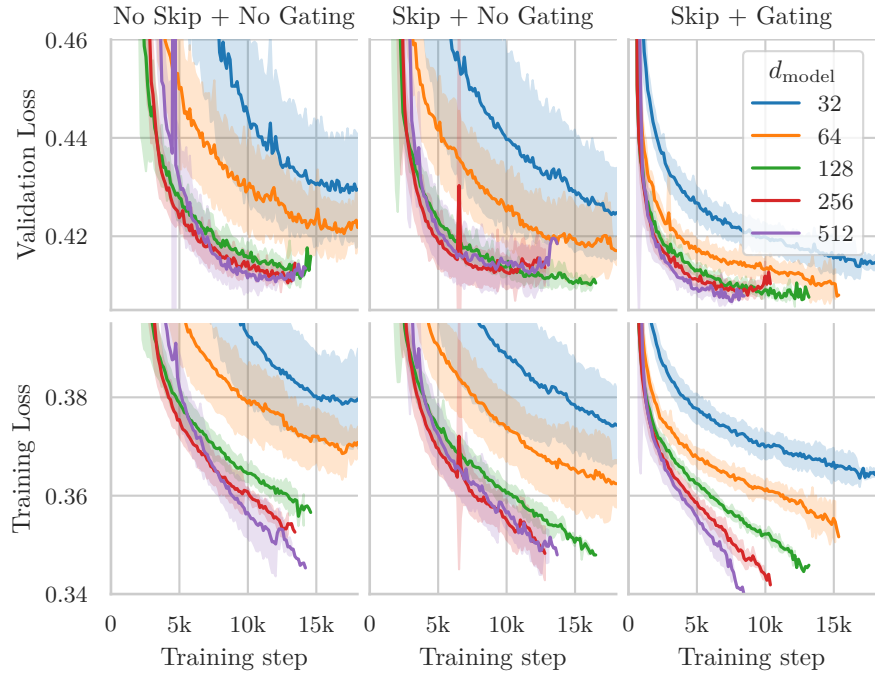}
  \caption{
    Validation and training loss on the M4-Monthly dataset.
    Each line represents the mean training loss over 9 repeated runs, with the shaded area representing the standard deviation.
    Please note that this plot contains a form of survival bias, as training is stopped once the validation loss flattens or increases.
  }
  \label{fig:skip_loss}
\end{figure}
\subsection{Ablation studies}
We perform two ablation studies, focusing only on the Monthly portion of M4, as it is the frequency with the most time series, contributing to 48\% of the overall dataset and total OWA score.
As mention in the previous section, we vary the model size by setting the hyper-parameter $d_\text{model}$ to values in the set $\{ 32, 64, 128, 256, 512 \}$, and set the feedforward size $d_\text{ff}$ to~$4\cdot d_\text{model}$.
For each combination of size setting and ablation variable setting, we train 9 models with different random weight initializations.

\subsubsection{Skip connection and multiplicative gating.}
The first ablation study focuses on the components of the proposed Persistence Initialization.
We compare an architecture with no skip connection and no multiplicative gating (Equation \ref{eq:skip_alpha_c}), an architecture with a skip connection and no multiplicative gating (Equation \ref{eq:skip_alpha_b}), and the full architecture with both a skip connection and multiplicative gating (Equation \ref{eq:skip_alpha_a}):

Figure \ref{fig:skip_owa_boxplot} shows box plots of the comparisons.
It is clear that the Transformer needs both the skip connection and the gating mechanism to improve as size increases.
Using no skip connection and no gating leads to worse performance for the largest model, while only using a skip connection leads to a flat trend with no performance increase as the model size increases.
Furthermore, even in the small setting of $d_\text{model}=64$, the models with both skip connection and gating mechanism clearly outperform the ones without a skip connection or a gating mechanism.

Training and validation loss is shown in Figure \ref{fig:skip_loss}.
The training loss shows clearly that the models with both skip connection and gating mechanism converge faster and to a lower loss value.
Moreover, the standard deviation of the losses is much smaller, indicating a more stable training process.

\begin{figure}[t]
  \centering
  \input{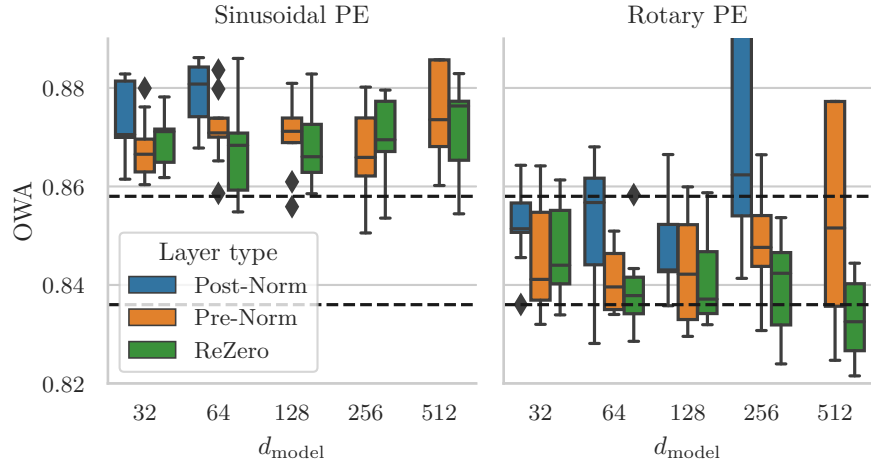}
  \caption{
    OWA test scores on the M4-Monthly dataset.
    The dashed lines correspond to the first and second place entry in the M4 competition.
    Each box represents 9 repeated runs.
  }
  \label{fig:ablation_boxplot}
\end{figure}
\subsubsection{Positional encoding and normalization.}
The second study compares the effect of the positional encoding and normalization layers.
We compare two kinds of positional encodings and three kinds of normalization layers, resulting in a total of six ablation combinations.
The two positional encodings are the standard sinusoidal positional encoding and the Rotary encoding.
The three kinds of normalization layers are post-activation Layer Norm, pre-activation Layer Norm, and ReZero normalization.
Figure \ref{fig:ablation_boxplot} shows box plots for the six ablation settings.
We observe that the Rotary positional encoding shows better performance in comparison to sinusoidal encoding, for all values of $d_\text{model}$ and for each of the three layer types.
Furthermore, the models with ReZero normalization show improved performance as the model size increases, in contrast to pre- and post-activation Layer Norm.

\subsection{Comparison with other Transformer models for Time Series}
\begin{table}[t]
  \centering
  \caption{
    Comparison of Transformer models on M4-Hourly.
    The score for the LogSparse Transformer is taken from \cite{li2019enhancing}, the others are median scores of 5 repeated experiments.
    $R_{0.5}$ is the median quantile loss, and is defined in Equation \ref{eq:normalized_deviation}.
  }
  \label{tab:M4_hourly}
  \begin{tabular}{lcc}
    \toprule
    {}                      & OWA            & $R_{0.5}$      \\
    \midrule
    LogSparse Transformer   & -              & 0.067          \\
    Informer                & 0.670          & 0.056          \\
    Autoformer              & 1.033          & 0.078          \\
    \textbf{PI-Transformer} & \textbf{0.525} & \textbf{0.046} \\
    \bottomrule
  \end{tabular}
\end{table}
We compare our proposed Transformer architecture against three other recent Transformers recently proposed for time series forecasting: LogSparse Transformer~\cite{li2019enhancing}, Informer~\cite{informer}, and Autoformer~\cite{wu2021autoformer}.
Table \ref{tab:M4_hourly} shows the results of the comparison.
As can be seen from the table, our method outperforms the others both in terms of OWA and in terms of $R_{0.5}$.

\section{Conclusion}
\label{sec:conclusion}
In this work, we presented Persistence Initialization, an adaptation of the Transformer architecture for time series forecasting.
We show that Persistence Initialization improves forecasting performance and reduces training time.

A single Transformer with Persistence Initialization is able to outperform the winner of the M4 competition on the M4 dataset.
Furthermore, using the mean prediction of only 9 Transformers achieves comparable performance to the current state art method, N-BEATS.

\bibliographystyle{splncs04}
\bibliography{bibliography.bib}

\begin{thebibliography}{10}
\providecommand{\url}[1]{\texttt{#1}}
\providecommand{\urlprefix}{URL }
\providecommand{\doi}[1]{https://doi.org/#1}

\bibitem{ba2016layer}
Ba, J.L., Kiros, J.R., Hinton, G.E.: Layer normalization. arXiv preprint
  arXiv:1607.06450  (2016)

\bibitem{bachlechner2021rezero}
Bachlechner, T., Majumder, B.P., Mao, H., Cottrell, G., McAuley, J.: Rezero is
  all you need: Fast convergence at large depth. In: Uncertainty in Artificial
  Intelligence. pp. 1352--1361. PMLR (2021)

\bibitem{chen2016xgboost}
Chen, T., Guestrin, C.: Xgboost: A scalable tree boosting system. In:
  Proceedings of the 22nd acm sigkdd international conference on knowledge
  discovery and data mining. pp. 785--794 (2016)

\bibitem{he2016deep}
He, K., Zhang, X., Ren, S., Sun, J.: Deep residual learning for image
  recognition. In: Proceedings of the IEEE conference on computer vision and
  pattern recognition. pp. 770--778 (2016)

\bibitem{hyndman2006another}
Hyndman, R.J., Koehler, A.B.: Another look at measures of forecast accuracy.
  International journal of forecasting  \textbf{22}(4),  679--688 (2006)

\bibitem{ioffe2015batch}
Ioffe, S., Szegedy, C.: Batch normalization: Accelerating deep network training
  by reducing internal covariate shift. In: International conference on machine
  learning. pp. 448--456. PMLR (2015)

\bibitem{li2019enhancing}
Li, S., Jin, X., Xuan, Y., Zhou, X., Chen, W., Wang, Y.X., Yan, X.: Enhancing
  the locality and breaking the memory bottleneck of transformer on time series
  forecasting. Advances in Neural Information Processing Systems  \textbf{32},
  5243--5253 (2019)

\bibitem{makridakis2020m4}
Makridakis, S., Spiliotis, E., Assimakopoulos, V.: The m4 competition: 100,000
  time series and 61 forecasting methods. International Journal of Forecasting
  \textbf{36}(1),  54--74 (2020)

\bibitem{montero2020fforma}
Montero-Manso, P., Athanasopoulos, G., Hyndman, R.J., Talagala, T.S.: Fforma:
  Feature-based forecast model averaging. International Journal of Forecasting
  \textbf{36}(1),  86--92 (2020)

\bibitem{oreshkin2019n}
Oreshkin, B.N., Carpov, D., Chapados, N., Bengio, Y.: N-beats: Neural basis
  expansion analysis for interpretable time series forecasting. arXiv preprint
  arXiv:1905.10437  (2019)

\bibitem{radford2019language}
Radford, A., Wu, J., Child, R., Luan, D., Amodei, D., Sutskever, I., et~al.:
  Language models are unsupervised multitask learners. OpenAI blog
  \textbf{1}(8), ~9 (2019)

\bibitem{smyl2020hybrid}
Smyl, S.: A hybrid method of exponential smoothing and recurrent neural
  networks for time series forecasting. International Journal of Forecasting
  \textbf{36}(1),  75--85 (2020)

\bibitem{srivastava2015highway}
Srivastava, R.K., Greff, K., Schmidhuber, J.: Highway networks. arXiv preprint
  arXiv:1505.00387  (2015)

\bibitem{su2021roformer}
Su, J., Lu, Y., Pan, S., Wen, B., Liu, Y.: Roformer: Enhanced transformer with
  rotary position embedding. arXiv preprint arXiv:2104.09864  (2021)

\bibitem{vaswani2017attention}
Vaswani, A., Shazeer, N., Parmar, N., Uszkoreit, J., Jones, L., Gomez, A.N.,
  Kaiser, {\L}., Polosukhin, I.: Attention is all you need. In: Advances in
  neural information processing systems. pp. 5998--6008 (2017)

\bibitem{wu2021autoformer}
Wu, H., Xu, J., Wang, J., Long, M.: Autoformer: Decomposition transformers with
  {Auto-Correlation} for long-term series forecasting. In: Advances in Neural
  Information Processing Systems (2021)

\bibitem{you2019large}
You, Y., Li, J., Reddi, S., Hseu, J., Kumar, S., Bhojanapalli, S., Song, X.,
  Demmel, J., Keutzer, K., Hsieh, C.J.: Large batch optimization for deep
  learning: Training bert in 76 minutes. arXiv preprint arXiv:1904.00962
  (2019)

\bibitem{zerveas2021transformer}
Zerveas, G., Jayaraman, S., Patel, D., Bhamidipaty, A., Eickhoff, C.: A
  transformer-based framework for multivariate time series representation
  learning. In: Proceedings of the 27th ACM SIGKDD Conference on Knowledge
  Discovery \& Data Mining. pp. 2114--2124 (2021)

\bibitem{informer}
Zhou, H., Zhang, S., Peng, J., Zhang, S., Li, J., Xiong, H., Zhang, W.:
  Informer: Beyond efficient transformer for long sequence time-series
  forecasting. In: Proceedings of AAAI (2021)

\end{thebibliography}

\end{document}